\documentclass[conference]{IEEEtran}
\usepackage{times}

\usepackage[sort&compress,numbers]{natbib}
\usepackage{multicol}
\usepackage[bookmarks=true,urlcolor=cyan,colorlinks=false]{hyperref}

\usepackage{amsfonts}
\usepackage{mathtools}
\usepackage{physics}
\usepackage{bm}
\usepackage{siunitx}

\usepackage{multirow}
\usepackage{booktabs}
\usepackage{makecell}

\usepackage{stfloats}

\usepackage{float}
\usepackage{cleveref}
\usepackage{graphicx}
\usepackage{subcaption}
\usepackage{url}

\newcommand\rurl[1]{%
  \href{http://#1}{\nolinkurl{#1}}%
}

\usepackage[nolist,nohyperlinks]{acronym}

\begin{acronym}
\acro{SDF}{Signed Distance Field}
\acro{TSDF}{Truncated Signed Distance Field}
\acro{RMP}{Riemannian Motion Policy}
\acrodefplural{RMP}{Riemannian Motion Policies}
\acro{LSTM}{Long Short-Term Memory}
\acro{FMM}{Fast Marching Method}
\acro{RNN}{Recurrent Neural Network}
\acrodefplural{RNN}{Recurrent Neural Networks}
\acro{FFN}{Feed-Forward Neural Network}
\acrodefplural{FFN}{Feed-Forward Neural Networks}
\acro{BCE}{Binary Cross Entropy Loss}
\acro{DAgger}{Data Aggregation}
\end{acronym}

\newcommand{\todo}[1]{}
\renewcommand{\todo}[1]{{\color{red} TODO: {#1}}}

\newcommand{\note}[1]{}
\renewcommand{\note}[1]{{\color{blue} NOTE: {#1}}}

\newcommand{\etal}{\textit{et al}.}

\DeclareMathOperator*{\argmax}{arg\,max}

\begin{document}

\title{Pushing the Limits of Reactive Planning:\\Learning to Escape Local Minima}

 \author{

\authorblockN{Isar Meijer, Michael Pantic, Helen Oleynikova, Roland Siegwart}
\IEEEauthorblockA{
{\tt\small{isarmeijer@gmail.com, mpantic@ethz.ch, helenoleynikova@gmail.com, rsiegwart@ethz.ch}}}
 \IEEEauthorblockA{All authors are with the Autonomous Systems Lab, ETH Zurich, 8092 Zurich, Switzerland.}
 
 }
\maketitle

\IEEEpeerreviewmaketitle

\begin{abstract}
When does a robot planner need a map? Reactive methods that use only the robot’s current sensor data and local information are fast and flexible, but prone to getting stuck in local minima. Is there a middle-ground between fully reactive methods and map-based path planners? In this paper, we investigate feed forward and recurrent networks to augment a purely reactive sensor-based planner, which should give the robot ``geometric intuition'' about how to escape local minima. We train on a large number of extremely cluttered worlds auto-generated from primitive shapes, and show that our system zero-shot transfers to real 3D man-made environments, and can handle up to 30\% sensor noise without degeneration of performance.  We also offer a discussion of what role network memory plays in our final system, and what insights can be drawn about the nature of reactive vs. map-based navigation. 
\end{abstract}

\section{Introduction}
Robots are increasingly being deployed in complex and cluttered environments. Collision-free navigation is one of the most fundamental and best-studied skills a robot must possess. Most collision avoidance methods fall into one of two categories:  map-based or reactive. \textit{Map-based} methods rely on a processed world representation which can be checked for collisions, while \textit{reactive} approaches often only use the robot's local information and current sensor data to decide the robot's next action.

Local reactive methods can be extremely computationally efficient and provide safety without relying on additional processes, such as mapping frameworks or state estimators. Additionally, strict assumptions about the environment, like static world assumptions, are not needed. However, without memory or a longer-term perspective they are prone to getting stuck in local minima -- geometric dead-ends such as large walls, U-shaped features, or long corridors. Ideally, even a purely reactive method would have a certain sense of geometric intuition that allows it to make informed decisions even in absence of a consistent map. How can we build such a system? Where are the limits of purely reactive navigation, in terms of geometric and temporal consistency? When is it better to rely on a map?

In this paper, we aim to answer these questions using a succession of novel methods for informed reactive navigation by combining a purely classical, reactive approach with different neural networks. We use simple \acp{FFN} and \acp{RNN} with \ac{LSTM} cells trained in a self-supervised environment to provide geometric ``intuition'' that is then combined with a classical safety layer. These networks essentially bias the classical reactive method to avoid and escape local minima.
Our method is made sensor-agnostic by expressing the sensor data as a set of \textit{rays} originating at the robot's current position.
\begin{figure}[t]
    \centering

    \includegraphics[trim={250px 100px 300px 0px}, clip, width=\linewidth]{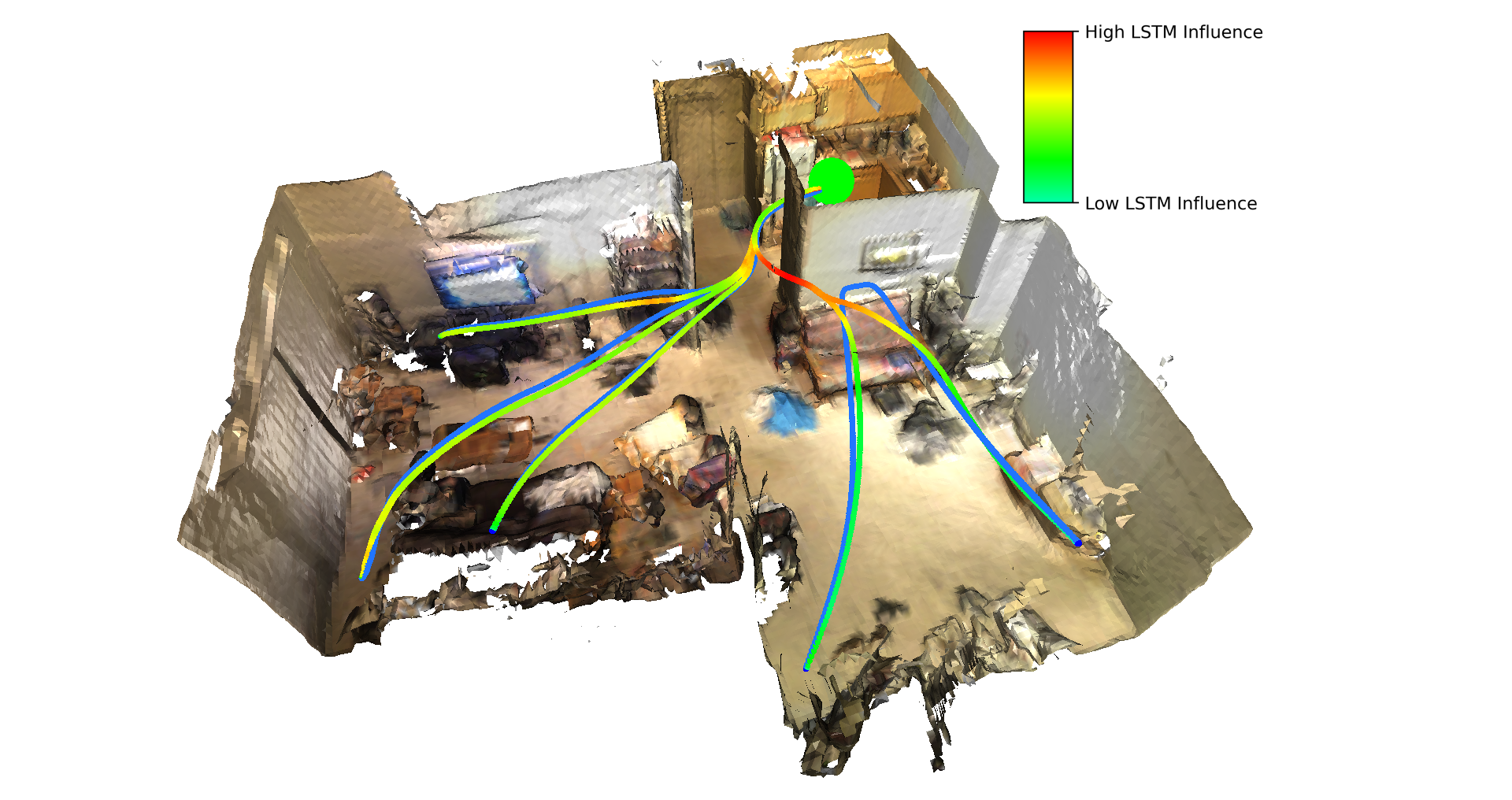}
    \caption{A comparison of baseline (blue) and learned \ac{LSTM} (rainbow) trajectories from various starting points to the goal location shown by a green sphere. The reactive baseline cannot navigate to the goal when it is directly behind a wall, while our learned method uses its ``geometric intuition'' (shown by the high influence of the \ac{LSTM}) to turn left to escape the local minimum. The world is from the BundleFusion dataset \cite{dai2017bundlefusion}, and the learned system is transferred zero-shot to this environment after training on synthetically generated data containing only obstacle primitives. 
    \vspace{-10pt}}
    \label{fig:teaser}
\end{figure}

By studying the performance implications of the different variants of our sensor-agnostic method, we provide novel insights into the limits of reactive navigation and to what extent these limitations can be overcome using different networks with varying degrees of temporal consistency. Most importantly, we provide insight into the \textit{nature} of the different methods -- we investigate how different parts of the network play a larger role on the resulting trajectory and what implications can be drawn from it.
Our method is loosely inspired by navigation of humans and animals, which are able to anticipate navigation decisions, without a metric or complete map, by using memory and intuition.
By training our approaches in auto-generated, unstructured environments with very high obstacle density (\Cref{fig:planning_examples}), many different shapes of local minima are encountered and learned. The resulting approaches are then evaluated on structured, human-made environments (\Cref{fig:teaser}).

The central goal of this work is to explore and understand how far the limits of reactive planning can be pushed without using an explicit map. To this end, we contribute:
\begin{itemize}
   \item multiple neural network architectures that provide ``geometric intuition'' to a purely reactive baseline planner,
   \item show how these networks generalize to real 3D environments with different levels of sensor noise,
   \item and a thorough discussion how temporal memory within the network affects the overall planning system's ability to escape local minima.
\end{itemize}
The purely reactive baseline method is reviewed in \Cref{sec:pure}, then combined with an additional feed-forward network in \Cref{sec:neural}, then extended to a recurrent network in \Cref{sec:recurrent}. For each method we discuss its performance, short-comings, and implications, based on qualitative examples. We then compare all variants with established methods that rely on a full map (i.e. are non-reactive) in \cref{sec:quant} and discuss the results and fundamental findings in detail in \Cref{sec:discussion}.
\section{Related Works}
Many works on robotic navigation rely on accurate maps of the environment. This allows the community to treat mapping and navigation as separate components with their own respective evaluation metrics.
The first works in this direction were \textit{potential field methods}~\cite{khatib1986potential}, which treated every obstacle as applying a repulsive force on the robot. Such a potential field was computed over the entire map, allowing the addition of other constraints or optimization objectives on the trajectory such as elastic band models~\cite{quinlan1993elastic} or explicit kinematic or dynamic constraints~\cite{ratliff2009CHOMP}.

However, even with full map knowledge, methods based on collision potentials get stuck in local minima when the scenes become too cluttered~\cite{oleynikova2018safe}, as they often result in non-convex optimization problems.
Other map-based methods such as RRT*~\cite{karaman2011sampling} trade-off computation time for asymptotic optimality and probabilistic completeness. Due to their runtime, they are often used for global planning in combination with a faster local planning layer, such as any of the map-based potential field methods.

Reactive, map-free navigation still uses this concept of obstacle-induced repulsive forces, as in \cite{montano1997real}. Newer approaches such as \acp{RMP}~\cite{ratliff2018riemannian} allow us to combine multiple ``local policies'', encoding different robot objectives based on local information, smoothly. Pantic \etal~\cite{pantic2023obstacle} formulate a pure avoidance policy with \acp{RMP} which allows fast, safe navigation without requiring any built maps.
However, their approach still suffers from occasionally getting trapped in local minima. We use the open source implementation of~\cite{pantic2023obstacle} as a baseline method. Mattamala \etal~\cite{mattamala2022efficient} use a geodesic field in a reactive navigation setup to avoid local minima, however, computing the geodesic distance field \textit{a priori} requires a consistent map of the environment.

Other approaches solve reactive navigation by learning an \textit{end-to-end} system: sensor data in, robot action out. This removes the need for an explicit map.
Some end-to-end methods end up learning the map \textit{implicitly} as part of the training process, by training in the same environment as the robot navigates in. For example, using only visual input to  navigate through 2.5D mazes~\cite{mirowski2017learning} or flying in 3D through cluttered environments~\cite{song2023learning}.

Other works focus on learning policies that work in a variety of environments. Ross \etal~\cite{ross2013learning} learn a mapping from image features to velocity commands from human pilot demonstration using \acs{DAgger}~\cite{ross2011reduction} for a drone flying rapidly through a forest; however, requiring human pilot demonstrations makes training for new environments difficult.
Loquercio \etal~\cite{loquercio2021learning} instead train fast dynamic drone flight on a variety of environments in simulation, and show impressive results of zero-shot transfer to the real world, using stereo disparity images as input and a short-horizon dynamic trajectory as output.
The main focus of their work is to fly \textit{quickly} through semi-cluttered environments, while we specifically want to focus on escaping local minima with a mix of classical and learned approaches. 

Finally, other end-to-end networks have used reinforcement learning with a similar ray-based sensor representation, but limited to 2/2.5D. Tai \etal~\cite{tai2017virtual} learn a general navigation policy in 2D from LiDAR scans which map to a velocity output, and generalize across sparse maps. Similarly, Pfeiffer \etal~\cite{pfeiffer2018reinforced} uses a similar set-up in 2.5D with imitating expert demonstrations. Zhang \etal~\cite{zhang2017neural} use explicit external memory structures to learn a representation of the visited environment, however, limited to discretized 2D worlds. 

By mixing a classical reactive avoidance algorithm with a supervised learning component, we are able to handle full 3D trajectories in vastly more cluttered environments than presented in other works. 

The goal of our work is to see how we can build on the ideas of \cite{pantic2023obstacle} to overcome the downsides of map-free, purely reactive navigation without requiring an explicit map, while leveraging the best advantages of both classical and learned methods.
We rely on the \ac{RMP}-based repulsive field approach to provide collision-free, safe trajectories, while our new learned component can focus on developing a geometric intuition about the environment.
The following sections will describe the purely reactive safety layer, followed by feed-forward and then recurrent network architectures.
\section{Purely Reactive Navigation}
\label{sec:pure}
We use the open-source system presented in \cite{pantic2023obstacle} as the base method for purely reactive navigation. Obstacle avoidance is formulated as a combination of obstacle repulsive forces, represented as \acp{RMP}~\cite{ratliff2018riemannian}, with each ``obstacle'' being generated by a ray-cast into a volumetric map \cite{millane2023nvblox}. Each ray creates one repulsion policy.
As it serves as our base method, and for the reader's convenience, we reproduce the most important concepts of \aclp{RMP} and how they are used to perform obstacle avoidance in this chapter. For more details we refer the reader to the respective original work \cite{ratliff2018riemannian, pantic2023obstacle}.
\subsection{Riemannian Motion Policies}
 An \ac{RMP} is a policy $\mathcal{P}$ that consists of an acceleration $\bm{f} \left(\bm{x}, \bm{\dot{x}}\right) \in \mathbb{R}^3$ coupled with a metric $\bm{A}\left(\bm{x}, \bm{\dot{x}}\right) \in \mathbb{R}^{3\times3}$, where $\bm{x}, \bm{\dot{x}} \in \mathbb{R}^3$ denote the robot's position and velocity.
 A single policy $\mathcal{P} = \left(\bm{f}, \bm{A} \right)$ describes an acceleration on the system, combined with a metric that captures the directional importance of that acceleration.
A set of policies $\left\{ \mathcal{P}_i \right\}$ can be summed into a single policy $\mathcal{P}_c$ by
\begin{equation}
\label{eq:RMPsum}
    \mathcal{P}_c = \sum_i \mathcal{P}_i = {\left( \left( \sum_i \bm{A}_i \right) ^+ \sum_i \bm{A}_i \bm{f}_i ,\, \sum_i \bm{A}_i \right)},
\end{equation}
where $^+$ denotes the pseudoinverse. The result is a policy itself, that contains an implicitly metric-optimal, joint behavior of all policies.
A policy can be a function of the robot's state, the goal location, and any other type of observation that is available. In the next section, we introduce local observations in the form of ray casts. 
\subsection{Ray-Casting}
\begin{figure}
    \begin{minipage}[t]{.55\linewidth}
    \centering
    \includegraphics[width=1.1\linewidth]{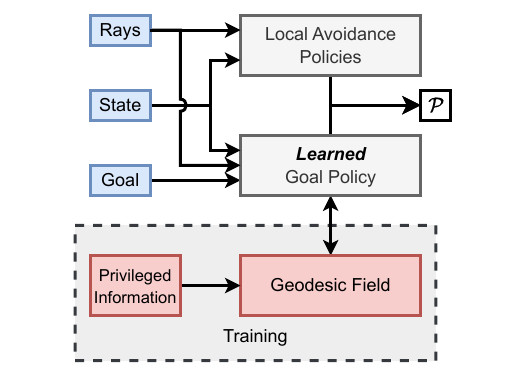}
    \captionsetup{width=.95\linewidth}
    \caption{System overview of the learned reactive planner. We combine the local avoidance policies as a safety layer with a higher level learned system, supervised by the geodesic field during training.\vspace{-10pt}}
    \label{fig:learned-reactive}
    \end{minipage}
    \begin{minipage}[t]{.44\linewidth}
    \centering
    \includegraphics[width=\linewidth]{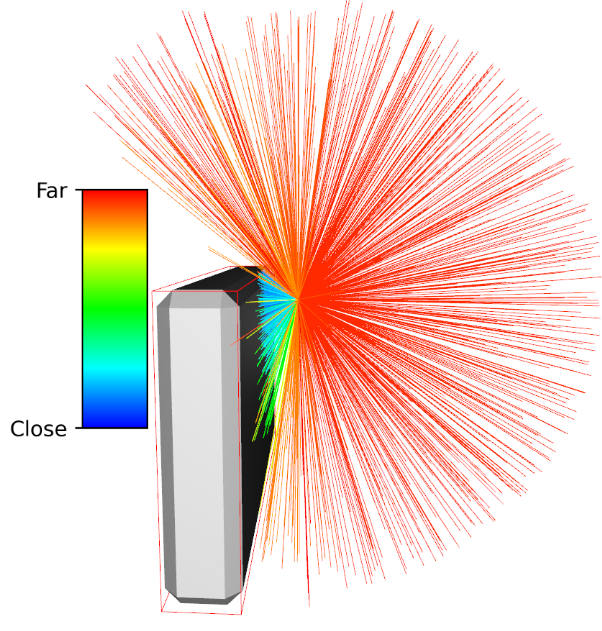}
    \captionsetup{width=.95\linewidth}
    \caption{Example of 1024 ray-casts obtained via Halton sampling. These rays are a map-agnostic method to gain local geometric information.\vspace{-10pt}}
    \label{fig:ray_example}
    \end{minipage}
    
\end{figure}

All following policies have access to state information and the goal location. Additionally, we use ray casts from the robot's current pose as a general abstraction of depth sensor data, as shown in \Cref{fig:ray_example}. 
For simplicity, we treat our robot as a point, but robot shape can be represented by modifying the distance of the rays; to model a sphere robot, a constant distance would be subtracted from all ray values for example. 

Similar to \cite{pantic2023obstacle}, we sample $N$ equally spaced, quasi-random depth rays by using the Halton sequence $\mathcal{H}\left(\cdot\right)$~\cite{halton1964algorithm}. We define the $i$'th Halton direction vector $\bm{r}_{\mathcal{H}}\left(i\right)$, parameterized by the elevation $\varphi_i$ and azimuth $\theta_i$, sampled from deterministic Halton sequences with base 2 and 3 respectively:
\begin{equation}
\begin{split}
\varphi_i &= \arccos\left(1 - 2 \cdot \mathcal{H}\left(i,\,2\right)\right), \\
\theta_i &= 2\pi \cdot \mathcal{H}\left(i,\,3\right),
\end{split}
\end{equation}
\begin{equation}
\bm{r}_\mathcal{H}\left(i\right) = \left(\sin{\varphi_i}\cos{\theta_i}, \cos{\varphi_i}\sin{\theta_i}, \cos{\varphi_i}\right).
\end{equation}
The ray obstacle distance function $d_r\left(\bm{x}, \bm{r}, L\right)$ returns the distance to an obstacle from the position $\bm{x}$ along a ray direction $\bm{r}$, truncated at a maximum distance $L$. Using this, we can define the ray-cast function $\bm{\mathcal{R}}^k : \mathbb{N} \times  \mathbb{R}_+ \rightarrow \mathbb{R}^N_+$ as
\begin{equation}
\label{eq:ray_cast}
    \bm{\mathcal{R}}^k \left(N, L\right) = \left[d_r\left(\bm{x}^k, \bm{r}_\mathcal{H}\left(i\right), L\right)\right]^{N - 1}_{i=0}.
\end{equation}
\Cref{eq:ray_cast} samples $N$ equally spaced rays $\bm{r}_\mathcal{H}\left(i\right)$ according to the ray-cast function from the robot's position $\bm{x}^k$ at time step $k$.
Due to the Halton sequence' deterministic nature, for function calls with identical $N$, the direction vector $\bm{r}_\mathcal{H}\left(i\right)$ for each element $i$ in the output remains the same across timesteps. By using a GPU-based mapping environment~\cite{millane2023nvblox}, this function can be evaluated in parallel~\cite{pantic2023obstacle} for all rays.
\subsection{Collision Avoidance}
\label{sec:reactive_obstacle_avoidance}
The obstacle avoidance policy defined in \cite{ratliff2018riemannian} serves as a building block for collision avoidance. The obstacle policy repels the robot from a point obstacle and consists of a repulsive term $\bm{f}_{rep}\left(\bm{x}\right)$ and a damping term $\bm{f}_{damp}\left(\bm{x},\, \bm{\dot{x}}\right)$:
\begin{equation}
    \bm{f}_{obs} \left( \bm{x}, \, \bm{\dot{x}} \right) = \bm{f}_{rep}\left(\bm{x}\right) + \bm{f}_{damp}\left(\bm{x},\, \bm{\dot{x}}\right).
\end{equation}
The repulsive term applies an acceleration away from obstacles based on distance, while the damping term is based on the velocity towards the obstacle. For more details and the definition of the corresponding metric, the reader is referred to the original work~\cite{ratliff2018riemannian}.
The combination of the repulsive term, damping term, and metric matrix results in a smooth, safe, and velocity-dependent avoidance behavior, where only obstacles that are very close to the robot or that the robot is about to approach have an effect. A simple unidirectional repulsor as used in potential field methods would react equally to all obstacles, and would not allow for parallel combination in the same fashion. This method allows us to combine thousands of such policies to perform avoidance in dense and  cluttered 3D maps. 
To do so, we follow the approach from \cite{pantic2023obstacle}, where the ray-cast function defined in \Cref{eq:ray_cast} is used to create an obstacle policy for every ray that hits an obstacle. A GPU implementation is leveraged such that not only ray-casting, but also policy creation and summation happens in parallel. 
\subsection{Goal Seeking}
\label{sec:reactive_goal_seeking}
Similar to \cite{pantic2023obstacle}, we use the goal policy as defined in \cite{ratliff2018riemannian} to implement goal seeking behavior. The goal policy pulls the robot towards a goal location $\bm{x}_g$ and is defined as 
\begin{equation}
\label{eq:target}
\begin{split}
    \bm{f}_g \left( \bm{x}, \bm{\dot{x}} \right) & = \alpha_g \bm{s} \left( \bm{x}_g - \bm{x} \right) - \beta_g \bm{\dot{x}}, \\
    \bm{A}_g \left( \bm{x}, \bm{\dot{x}} \right) & = \mathbb{I}^{3 \times 3} ,
\end{split}
\end{equation}
where $\alpha_g$ and $\beta_g$ are scalar tuning parameters, and $\bm{s}$ is a soft-normalization function~
\cite{ratliff2018riemannian}.
The robot is then commanded with the sum of all activated policies according to \Cref{eq:RMPsum}. Trajectories are generated by numerical integration of the resulting acceleration, starting from an initial position at rest. 
\subsection{Limitations}
This purely reactive method performs surprisingly well even in very cluttered and complex 3D maps (see also \cite{pantic2023obstacle}). Its navigation capabilities emerge from the sum of many simple obstacle avoidance policies and the goal seeking behaviour, but it can get trapped in local minima whenever the two parts cancel each other out.
\begin{figure}[t!]
    \includegraphics[trim={200px 150px 200px 100px}, clip, width=.95\linewidth]{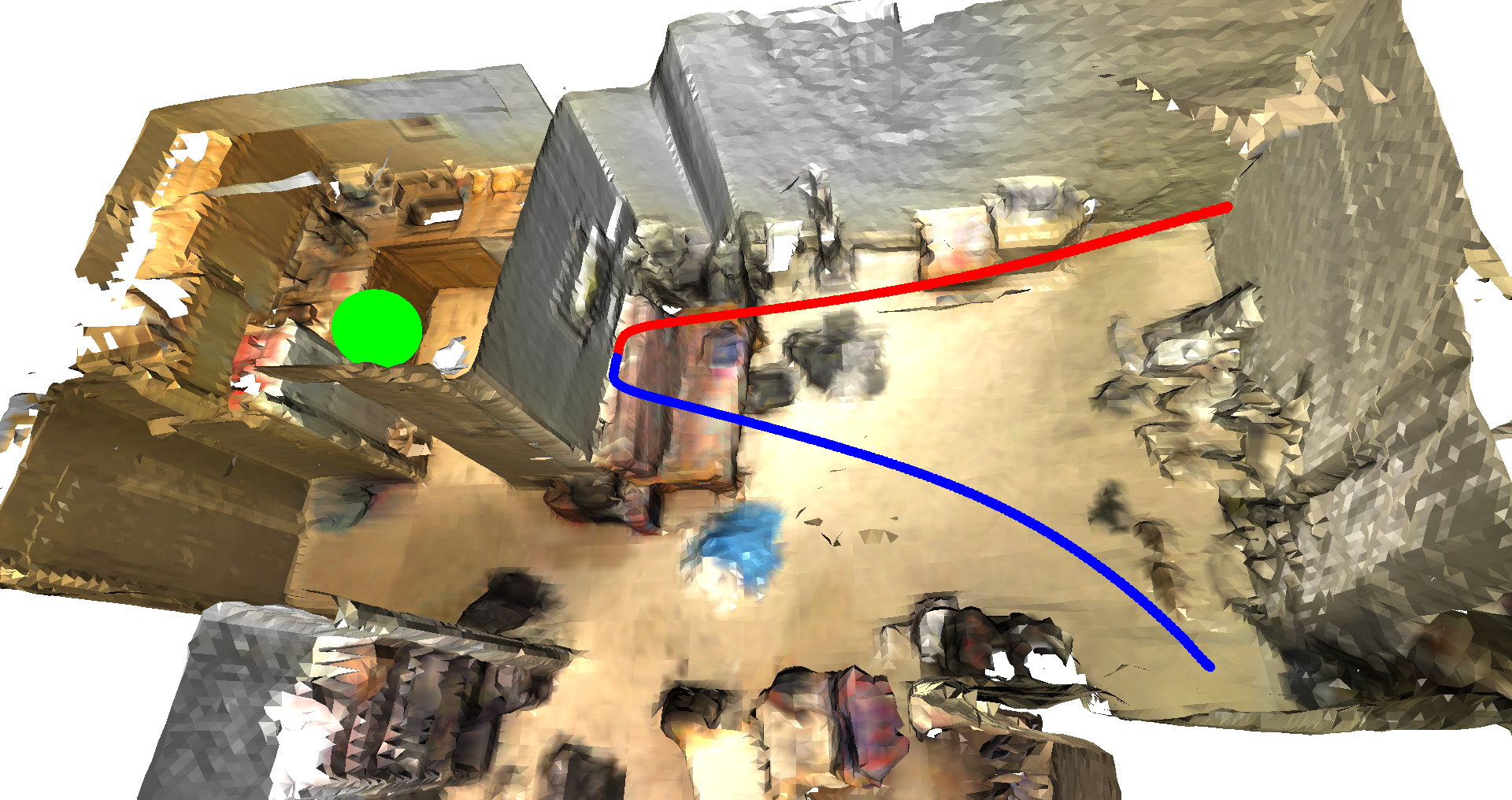}
    \caption{Example of 2 purely reactive planning runs getting stuck in the same local minimum. The goal location is behind the wall, depicted in green. The world is from the BundleFusion dataset \cite{dai2017bundlefusion}.\vspace{-10pt}}
    \label{fig:local_minima}
\end{figure}
Both, natural and man-made environments occasionally contain such local minima. \Cref{fig:local_minima} visualizes a typical example. However, humans are able to navigate these environments without a full map and do not get stuck indefinitely. Clearly, there are heuristics and intuitions about taking decisions that avoid or escape local minima. This naturally raises the question of how such intuition can be provided to a robotic system, preferably through a self-emergent process. In the next section, we introduce such an add-on to the purely reactive method by using a neural network that is trained in a self-supervised fashion.
\section{Neural Reactive Navigation}
\label{sec:neural}
We regress a geometric-aware informed goal policy as a function of the robo-centric goal direction and the sensor rays. The sensor rays provide the network with a sampling of the local geometry.
As a training signal, we use the \textit{geodesic distance field}. The geodesic distance field is a global function that captures the shortest distance to the goal from anywhere in the world around obstacles. It can be computed using the \ac{FMM}~\cite{sethian1996fast}, a grid-based version of which is implemented in the \textit{scikit-fmm}\footnote{\rurl{github.com/scikit-fmm/scikit-fmm}} library. The geodesic distance field requires perfect world information and is expensive to compute, therefore we only use it as privileged information for self-supervision during training. Doing so, the network should learn to infer near-optimal decisions that mimic the geodesic distance field directly from raw sensor data, effectively obtaining good heuristics to deal with local geometry.
A high-level overview of the proposed system is shown in Figure \ref{fig:learned-reactive}.

More formally, let the (symmetric) scalar function $\mathcal{G}\left(\bm{x}, \bm{x}_g\right)$ denote the geodesic distance field between locations $\bm{x}$ and $\bm{x}_g$. The shortest direction to the goal $\bm{x}_g$ from location $\bm{x}$ can be calculated by taking the (negative) gradient of $\mathcal{G}$ with respect to $\bm{x}$: 
\begin{equation}
\label{eq:geodesic}
    -\nabla_{\bm{x}}\, \mathcal{G}\left(\bm{x}, \bm{x}_g\right) .
\end{equation}
With this improved goal direction, we can create a goal policy similar to \Cref{eq:target}:
\begin{equation}
\label{eq:geodesic_target}
    \bm{f}_g \left( \bm{x}, \bm{\dot{x}} \right) = \alpha_g \bm{s} \left( -\norm{\bm{x}_g - \bm{x}}\nabla_{\bm{x}}\, \mathcal{G}\left(\bm{x}, \bm{x}_g\right)\right) - \beta_g \bm{\dot{x}},
\end{equation}
where we have changed the direction to the goal into the negative geodesic distance field gradient, while maintaining the original distance to the goal. The aim is to learn a function $\phi_\theta$ parameterized by $\theta$ that mimics \Cref{eq:geodesic}, using only the relative goal location and sensor rays as input.

\subsection{Architecture and Training}
\begin{figure*}
    \centering
    \includegraphics[width=.9\textwidth]{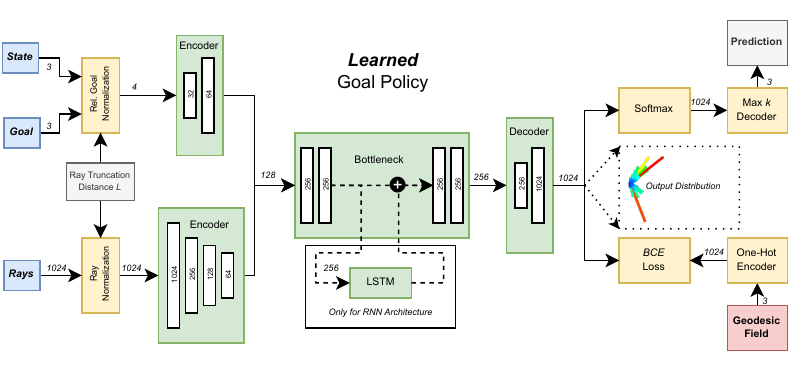}
    \caption{Network architecture. Learned blocks are highlighted in green. All learned blocks consist of a linear layer followed by layer normalization and a Leaky ReLU, except for the final decoder layer, which contains only a single linear layer. Layer normalization is used over alternatives like batch normalization, as it is more compatible with sequenced data. The \ac{LSTM} is only included in the recurrent neural reactive navigation system.
    \vspace{-10pt}}
    \label{fig:system_overview}
\end{figure*}

We regress the function $\phi_\theta$ using a multi-layer perceptron trained in a self-supervised fashion. Figure \ref{fig:system_overview} illustrates the full network architecture. In this section, we discuss architectural details, data generation and training methods. 

\subsubsection{Encoders}
The encoder part of the network consists of the ray encoder and the goal direction encoder. All input signals are rescaled to relative quantities with respect to the maximum ray length $L$.
The rays are linearly rescaled with respect to $L$. 
The state and goal information are used to calculate the relative goal direction $\bm{d}_g^k = \bm{x}_g - \bm{x}^k$, which is fed to the network as a unit direction $\bm{\hat{d}}_g^k = \bm{d}_g^k \big/ \norm{\bm{d}_g^k}$ vector and a scalar distance measure:
\begin{equation}
\begin{bmatrix}
    \bm{\hat{d}}_g^k \vspace{5pt}\\ 
    \mathcal{D}\left(\norm{\bm{d}_g^k}, L\right)
\end{bmatrix} 
\in \mathbb{R}^4 , 
\end{equation}
with the distance normalization function $\mathcal{D}$ as 
\begin{equation}
    \mathcal{D}\left(d, L\right) = 
    \begin{cases}
        \frac{d} {2L}, & d \leq L \\
        \sigma\left(2 \frac{d - L}{L}\right), & d > L
    \end{cases},
\end{equation}
where $\sigma$ is the sigmoid function. This function maintains linearity below the ray truncation distance and saturates large input values to $1$. Doing so, the distance to the goal becomes a relative quantity with respect to the ray lengths, which also implicitly encodes if the goal location is in front or behind a perceived obstacle. 
\subsubsection{Information Bottleneck}
The concatenated state and ray latent representations are passed through four fully connected layers. 
\subsubsection{Decoder}
\label{sec:ffn_decoder}
Instead of directly regressing the geodesic gradient, we decode the latent space into a weighted output distribution of directional rays, similar to as they are used in the input rays. For calculating the loss from the geodesic distance field that fits this output, the label $\bm{y}$ is encoded as a one-hot vector $\bm{e}_i$ with index $i$ determined as,
\begin{equation}
    \argmax_i{\left\{  \bm{y}^T \bm{r}_{\mathcal{H}}\left(i\right) \right\}} ,
\end{equation}
i.e. we select the index of the Halton direction vector that is closest in direction to the label.

We train this network using a \ac{BCE} loss between the one-hot encoded label and the output of the decoder. While the supervision signal may be an exclusive class label, interpreting the outputs as independent binary classifiers using \ac{BCE} gives the model more freedom to encode multi-modal outputs. An additional benefit of using such an output format is that it enables easy introspection by visualizing the output distribution.

To obtain an \ac{RMP} compatible acceleration vector from the output distribution, we pass it through a softmax layer and calculate a weighted sum of the direction vectors, over the max $k$ largest outputs:
\begin{equation}
\label{eq:max_k_output_decoder}
    \bm{\hat{y}} = \sum_{i \in \mathcal{Y}^k} \bm{e}_i^T s\left(\bm{\hat{y}}_{\mathcal{H}}\right)  \bm{r}_\mathcal{H}\left(i\right) ,
\end{equation}
where $\mathcal{Y}^k$ is the index set of the largest $k$ output entries of $\bm{\hat{y}}_\mathcal{H}$, $\bm{e}_i$ is the $i$'th standard vector, and $s\left(\cdot\right)$ is the softmax function. Higher values of $k$ lead to smoother trajectories, however, we found that very high values somewhat decrease success rates. In all following experiments, the value of $k$ is set at $50$, which we found to provide a good trade-off between smoothness and success rates. 
\begin{figure}
  \centering
  \begin{subfigure}[t]{.25\textwidth}
    \centering
    \captionsetup{width=.9\linewidth}%
    \includegraphics[width=.98\linewidth]{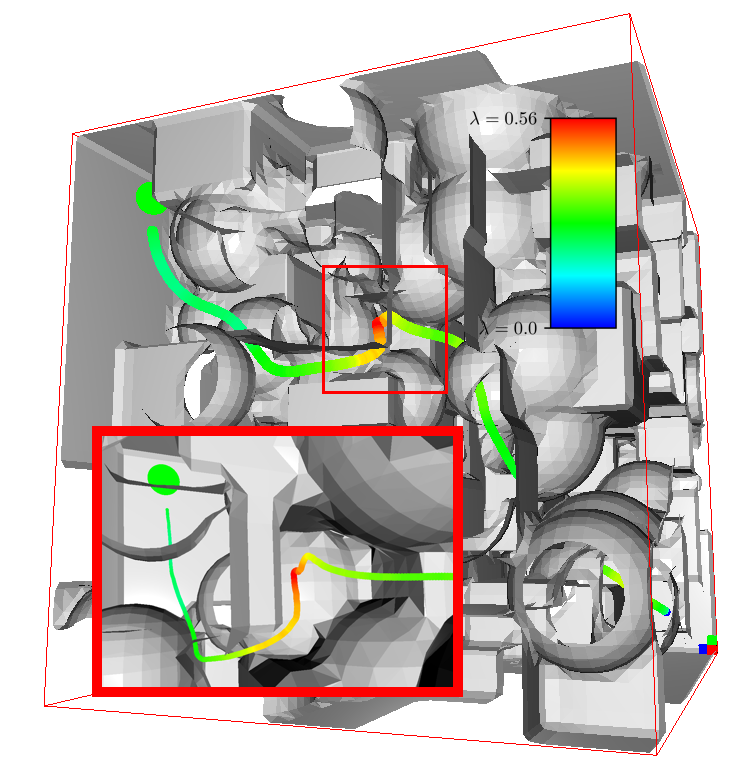}
    \caption{Sphere box world, 200 obstacles}
    \label{fig:sphere_box_planning_example}
  \end{subfigure}%
  \begin{subfigure}[t]{.25\textwidth}
    \centering
    \captionsetup{width=.9\linewidth}
    \includegraphics[width=.98\linewidth]{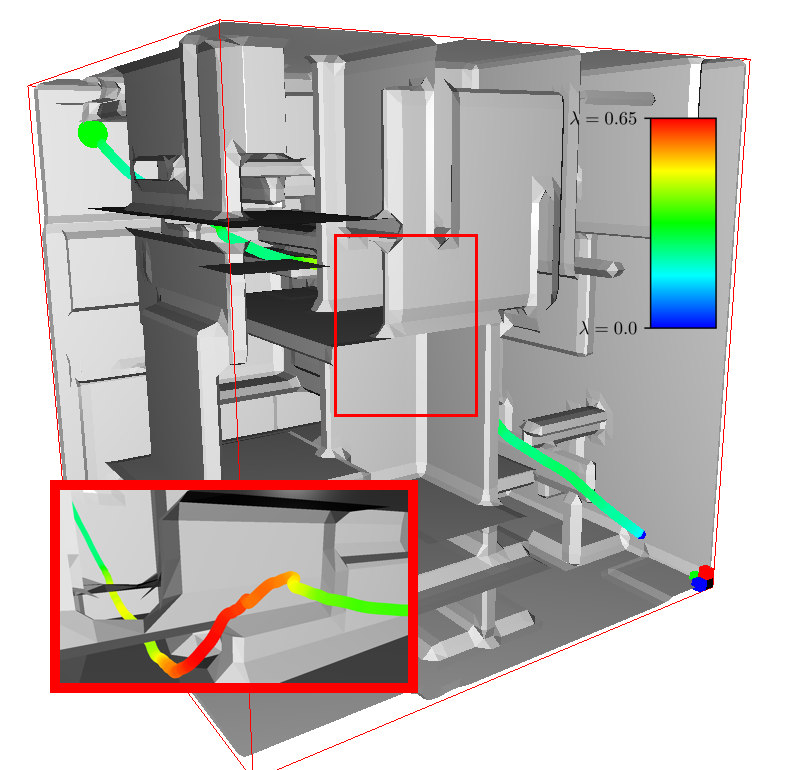}
    \caption{Plane world, 100 obstacles}
    \label{fig:plane_planning_example}
  \end{subfigure}%
  \caption{Examples of different planning runs in \textit{sphere box} and \textit{plane} worlds. The start and goal locations are depicted with the blue and green spheres respectively. The color gradient encodes the relative influence level of the \ac{LSTM} on the latent state. Note that the zoomed in part in the red rectangle has a different viewpoint compared to the outer view. Both these figures show emerging exploration behavior in the \ac{RNN} when navigating close to local minima.
  \vspace{-10pt}}
  \label{fig:planning_examples}
\end{figure}
\subsubsection{Training}
\label{sec:training}
We exclusively train on auto-generated worlds obtained through boolean combinations of primitive shapes. Two classes of $10\times10\times10\SI{}{\metre^3}$ worlds are created and used during training, validation, and testing: \textit{sphere box} worlds (\Cref{fig:sphere_box_planning_example}) and \textit{plane} worlds (\Cref{fig:plane_planning_example}). We use a varying amount of obstacles for each; up to $200$ obstacles for the \textit{sphere box} worlds, and up to $100$ obstacles for the \textit{plane} worlds. These classes of worlds are simple to generate in thousands of variations, without the need for human labeling or manual data collection. But more importantly, they resemble human-made structures in terms of local minima and entropy, which should facilitate generalization to unseen, structured environments. 
We train the network by using densely sampled random data from $6400$ of these randomly generated worlds. A random goal location is sampled for each, and subsequently $1024$ different locations are sampled within the world, from which the inputs and geodesic field label are collected. 
\subsection{Evaluation and Limitations}
\begin{figure*}[ht]
  \centering
  \begin{subfigure}[t]{.25\linewidth}
    \centering
    \captionsetup{width=.95\linewidth}%
    \includegraphics[trim={250px 150px 300px 100px}, clip, width=.98\linewidth]%
    {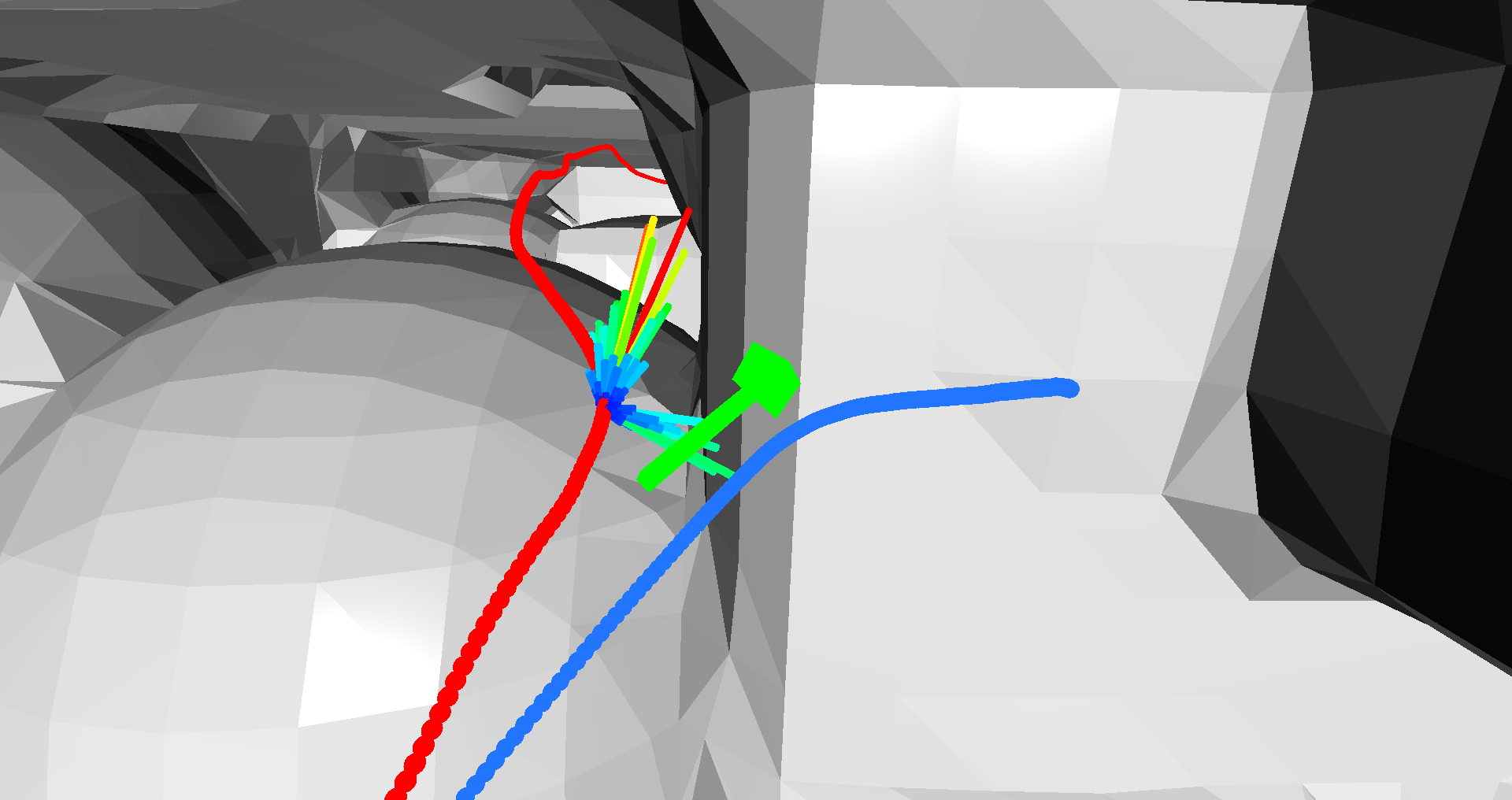}
    \caption{Baseline vs. \ac{FFN}. }
    \label{fig:ffn_vs_baseline}
  \end{subfigure}%
  \begin{subfigure}[t]{.25\linewidth}
    \centering 
    \captionsetup{width=.95\linewidth}
    \includegraphics[trim={250px 150px 300px 100px}, clip, width=.98\linewidth]%
    {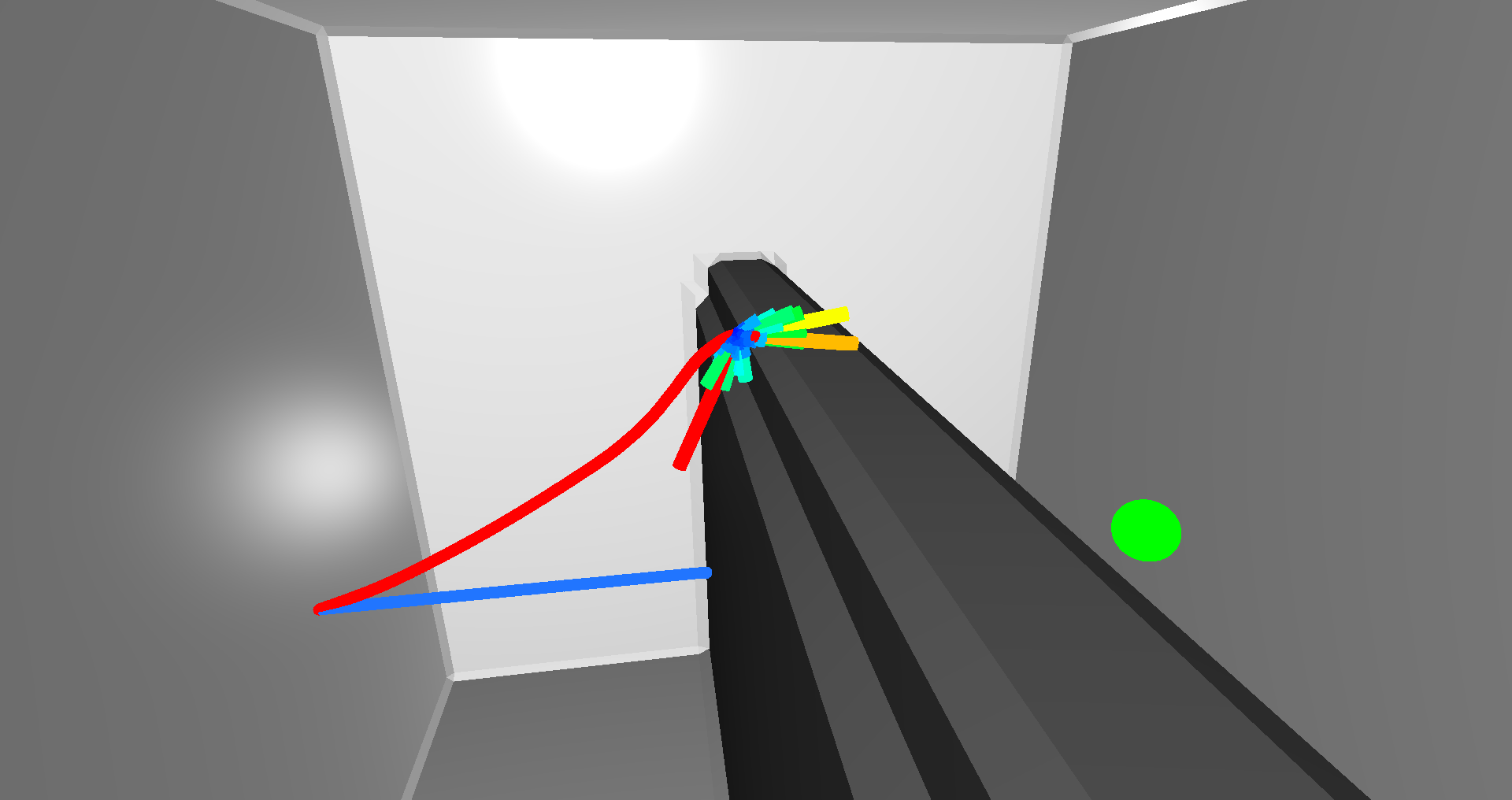}
    \caption[width=.95\linewidth]{Baseline vs. \ac{FFN}.}
    \label{fig:ffn_stuck}
  \end{subfigure}%
  \begin{subfigure}[t]{.25\linewidth}
    \centering
    \captionsetup{width=.95\linewidth}
    \includegraphics[trim={250px 150px 300px 100px}, clip, width=.98\linewidth]%
    {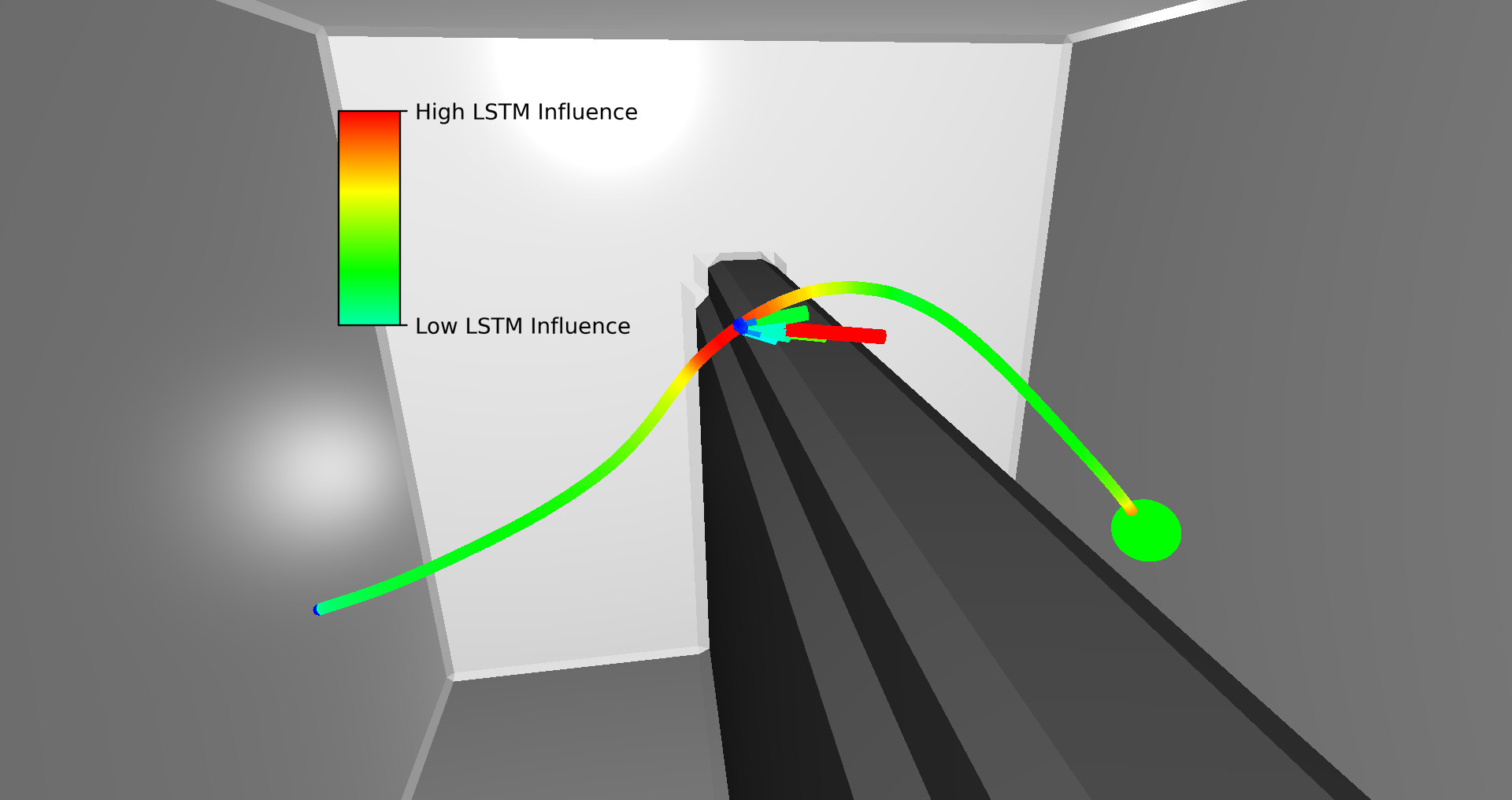}
    \caption{\ac{RNN}. }
    \label{fig:rnn_over_wall}
  \end{subfigure}%
  \begin{subfigure}[t]{.25\linewidth}
    \centering
    \captionsetup{width=.95\linewidth}
    \includegraphics[trim={150px 0px 400px 250px}, clip, width=.98\linewidth]%
    {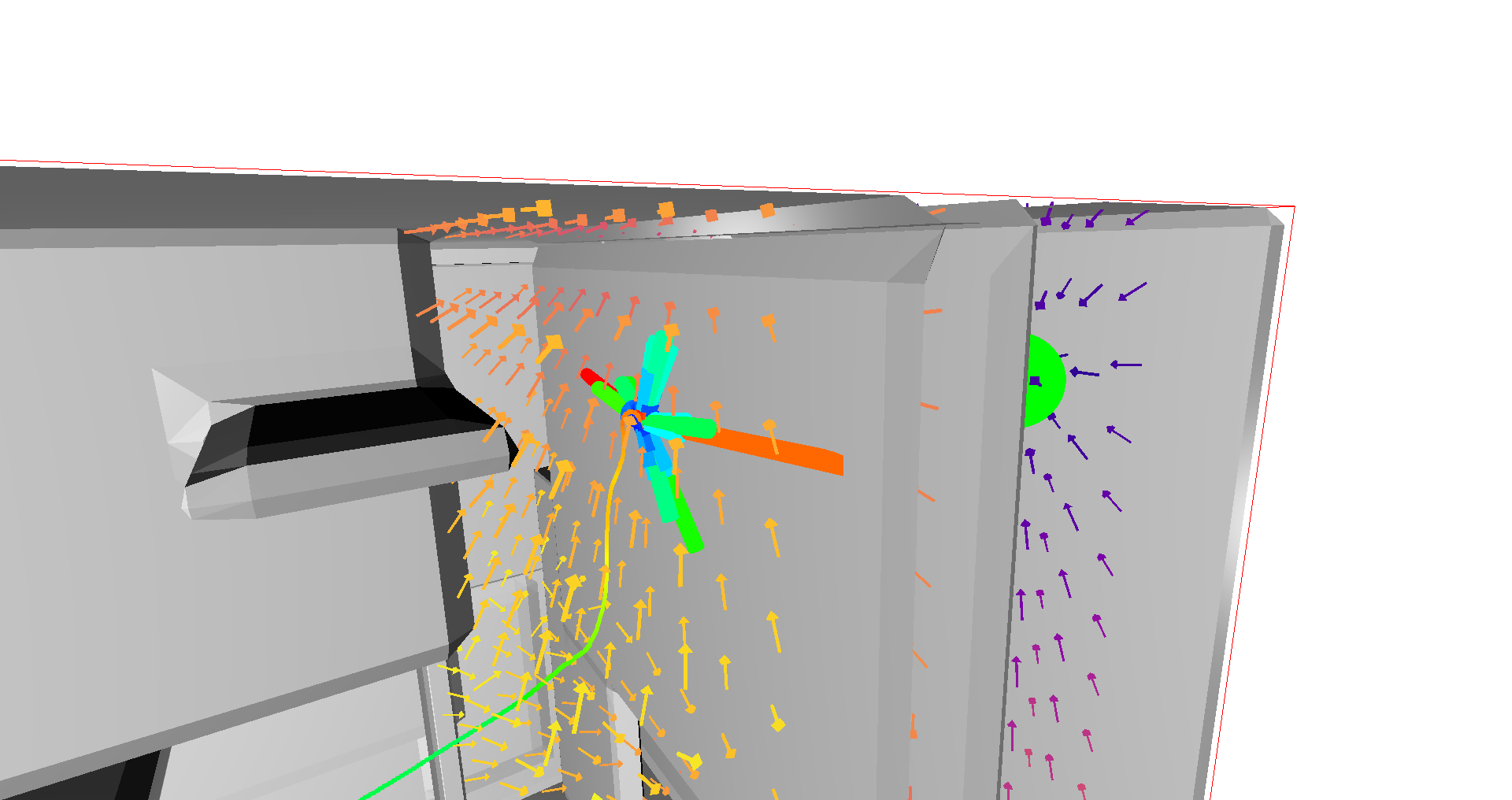}
    \caption{\ac{RNN}.}
    \label{fig:rnn_stuck}
  \end{subfigure}
  \caption{Examples of different planning runs, where the start and goal location are depicted by the the blue and green spheres respectively (in (a) the green arrow is pointing in the direction of the goal). The baseline is depicted in blue, the FFN is depicted in red, and the RNN is depicted with a color gradient depicting the influence of the \ac{LSTM}. We also show the output probability distribution of the learned systems at certain locations in the trajectory, where the color gradient and the length of the output `rays' are both a measure for the predicted output probability in that direction. In (a) the \ac{FFN} navigates past a local minimum in which the baseline gets stuck. In (b) the \ac{FFN} gets stuck on the corner of a wall as only having access to the current observation provides insufficient information on where to go, highlighted by the bimodal output distribution. In (c) the \ac{RNN} is able to navigate around the same wall, and we show that the LSTM has the highest influence exactly where the \ac{FFN} got stuck, causing the collapse of the second output mode observed in the \ac{FFN}. In (d) the \ac{RNN} gets stuck in a large local minimum, and the model has a large number of modes in the output distribution. The arrows depict the gradient of the geodesic field, and are colored by the geodesic distance to the goal.\vspace{-10pt}}
  \label{fig:overall}
\end{figure*}
The learned method outperforms both the baseline method and the comparison method (CHOMP~\cite{ratliff2009CHOMP}) significantly, as shown in \Cref{fig:success_rates}. The network is able to provide geometric intelligence to avoid local minima, while the baseline method still serves as a safety layer. The learned model implicitly acts in the sort of ``null-space" occurring whenever the obstacle avoidance potentials are not overly strong. 

While the system performs much better than the baseline, there are still obvious cases where it gets stuck, as seen for example in \Cref{fig:ffn_stuck}: local observations may provide insufficient information about the free space around the perceived obstacle, confirmed by the bimodal output distribution of the model. These cases can be attributed to the purely reactive nature of even the learned method. In some situations, regardless of the sensor used, previously observed data is needed to form a consistent model of the geometry in order to navigate it.
Those cases give rise to the need of temporal consistency -- which we will explore in the next section by using recurrent neural networks.
\section{Recurrent Neural Reactive Navigation}
\label{sec:recurrent}
We introduce memory by inserting a single \ac{LSTM} layer into the bottleneck, as depicted in \Cref{fig:system_overview}. The previous pure-FFN architecture was trained using randomly sampled dense data, an \ac{RNN} however requires sequences to train the recurrent elements. Therefore, we train the network by using \acf{DAgger}~\cite{ross2011reduction}; model rollouts from a random start and goal location are aggregated with previously collected runs during training. We found that the densely sampled dataset, however, proved to be more efficient at training a pure FFN model than using \ac{DAgger}. To combine the apparent training efficiency of the densely trained FFN, and the expected performance boost of including an \ac{LSTM}, we utilize a 2-stage training setup. First, the \ac{FFN} architecture is trained as described in Section \ref{sec:neural}. All weights are then frozen, the \ac{LSTM} is inserted and subsequently trained using \ac{DAgger}. 
\subsection{Evaluation and Limitations}
We evaluate the training performance of our recurrent models by directly comparing against a pure-FFN architecture: During training, a \ac{DAgger} validation dataset $\bm{D}_{\text{DAgger}}$ is collected, and both the model under training $\mathcal{M}$, and a benchmark pure-FFN model $\mathcal{M}_{B}$ trained previously are evaluated on that dataset using the loss function $\mathcal{L}$. The pure-FFN model is not influenced by temporal dependencies in the data, and is thus able to act as a reference for evaluation. We use the ratio $\mathcal{L}\left(\mathcal{M}\left(\bm{D}_{\text{DAgger}}\right)\right) \big/ \mathcal{L}\left(\mathcal{M}_B\left(\bm{D}_{\text{DAgger}}\right)\right) $ as a metric to compare different recurrent models. 
We use the learned model from \Cref{sec:neural} as the reference model. 

In \Cref{fig:dagger_loss_ratio} we compare this metric during training for the \ac{FFN}, \ac{RNN} and pretrained \ac{RNN} with frozen weights. We can see that the addition of the LSTM cell provides a performance increase over the \ac{FFN} when both are randomly initialized, however, a pretrained model with frozen weights is required to achieve a loss ratio below $1.0$. We speculate that the highly correlated data points within a trajectory sequence and the lack of data diversity due to rollouts naturally staying further away from obstacles contribute to the overall worse performance of only using \ac{DAgger} for the full network.

The \ac{LSTM}-based navigation system learns to overcome local minima for which using only immediate sensor data is insufficient.  We obtain qualitative evidence of the \acp{LSTM} capability of providing temporal consistency by coloring the trajectory based on the relative influence of the \ac{LSTM} on the latent representation inside the network. \Cref{fig:rnn_over_wall} shows a situation in which the LSTM is able to go over a wall in which the previous systems got stuck (see also \Cref{fig:ffn_stuck}). We observe only a single mode remains in the output distribution, and that the \ac{LSTM} has more influence in the exact locations where the \ac{FFN}-variant got stuck.
\Cref{fig:sphere_box_planning_example} visualizes the \ac{LSTM} influence on difficult randomly generated synthetic maps. We also evaluate the RNN on human made environments, shown in \Cref{fig:teaser} and \Cref{fig:real_world}. The learned system is able to generalize zero-shot to these structured environments from training entirely on synthetic datasets made from primitive shapes. We see that the \ac{LSTM} increases its influence at key locations for navigation in both the synthetic and real world examples.

As with any method, there are still limitations. Local minima that are considerably larger or deeper than observed on training data can still pose a problem, examples are visualized in \Cref{fig:rnn_stuck} and \Cref{fig:real_world_bundle_apt0}. This can be attributed to \textit{(a)} the training environments and \textit{(b)} the multi-modality of the output. Training on potentially synthetic but human-made structured environments is a promising further avenue for improvements. Additionally, choosing a more complex strategy to select the most promising ``mode'' of the multi-modal output than the weighted sum presented in \Cref{sec:ffn_decoder} could also improve the influence of the learned goal policy. \Cref{fig:rnn_stuck} depicts an example of a case where an improved strategy would be beneficial.
\section{Quantitative Evaluation}
\label{sec:quant}
In this section, we will quantitatively compare our proposed methods to existing solutions, both on synthetic and real-world scenes.

\subsection{Synthetic Scenes}
\begin{figure*}[ht!]
\begin{minipage}[t]{.33\linewidth}
    \centering
    \includegraphics[width=\linewidth]{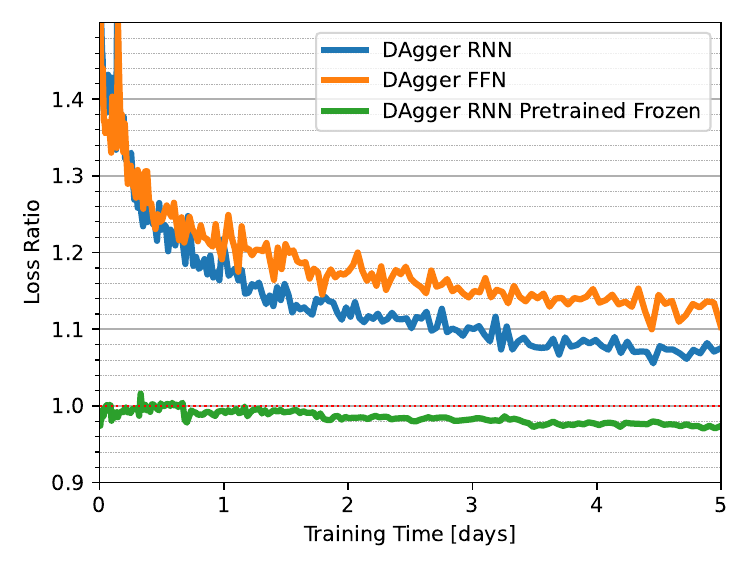}
    \captionsetup{width=\linewidth}
    \caption{Relative validation loss of different DAgger training runs w.r.t. the reference \ac{FFN} model. The randomly initialized \ac{FFN} and \ac{RNN} show that the addition of the LSTM provides a performance boost. However, pretraining and freezing the non-LSTM parts of the network is required to achieve the best performance.\vspace{-10pt}}
    \label{fig:dagger_loss_ratio}
\end{minipage}
\begin{minipage}[t]{.66\linewidth}
    \centering
    \includegraphics[width=\linewidth]{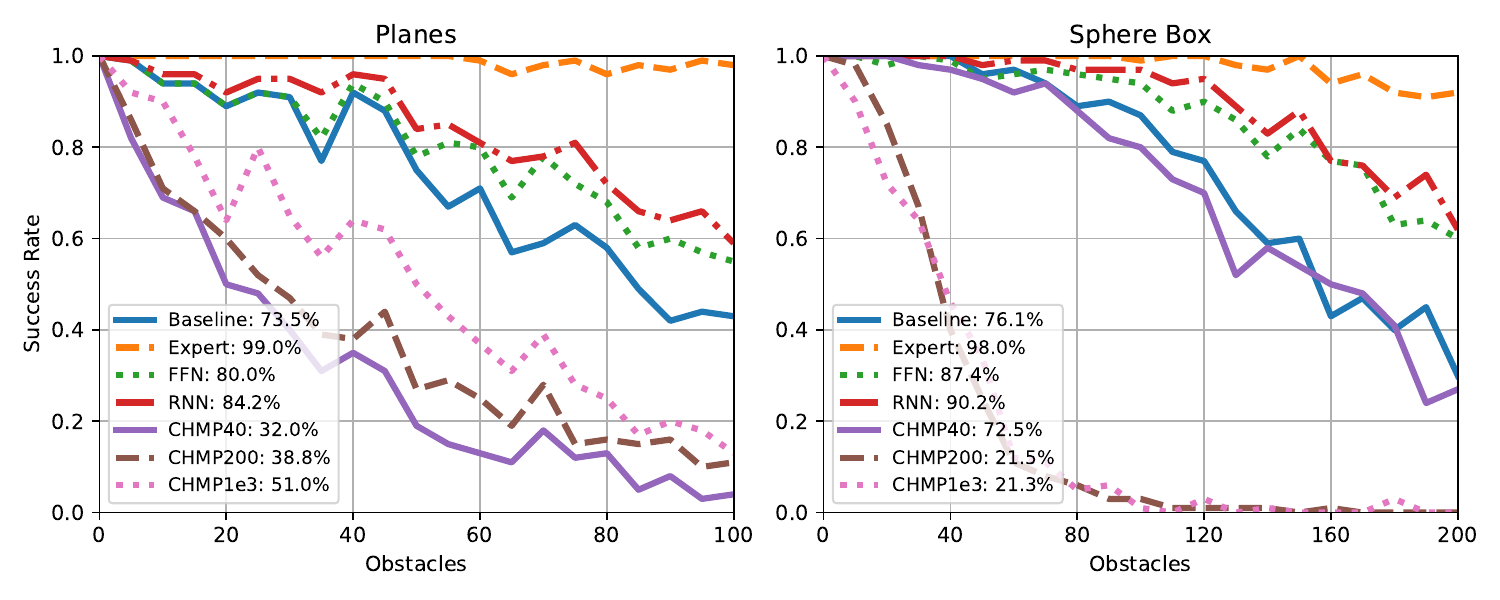}
    \captionsetup{width=.9\linewidth}
    \caption{Success rates as a function of obstacle density. The percentages in the legend denote the overall success rates. $N = 2100$ per world type and algorithm (100 runs per datapoint). The reactive planners outperform all configurations of the map-based CHOMP, even though the reactive methods do not have access to the map. Both learned systems outperform the baseline, with the \ac{RNN} also slightly outperforming the \ac{FFN}.\vspace{-10pt}}
    \label{fig:success_rates}
\end{minipage}
\end{figure*}

To evaluate how our reactive method, that uses only current sensor data, compares to methods that use a complete map, we compare against an ``expert'', which simply follows the geodesic field gradient, and CHOMP, a well-known local planner~\cite{ratliff2009CHOMP}.
We compare planning success rates as a function of obstacle density on a separate test set, depicted in \Cref{fig:success_rates}. In all following evaluations, ``baseline'' refers to the system as described in \cref{sec:pure}, ``\ac{FFN}'' to the one in \cref{sec:neural}, and ``\ac{RNN}'' to \cref{sec:recurrent}. ``CHMP $c$'' corresponds to CHOMP~\cite{ratliff2009CHOMP} with a collision weight $c$, and ``Expert'' is a policy that directly uses the gradient of the geodesic distance field as input, as shown in \Cref{eq:geodesic_target}. The expert does not always achieve a $100\%$ success rate, as it is still part of a policy-based system; i.e. in certain cases the local avoidance policies do not let the expert pass through very narrow spaces. 
It is important to note that CHOMP and the expert both use \textbf{privileged} information (the full map), whereas all other systems \textbf{only} have access to immediate sensor rays.

The \ac{FFN}- and \ac{RNN}-based systems outperform all other local methods for almost all map variants. It seems that despite having access to the full map, CHOMP struggles with the quasi-global planning problems we used here to push local navigation algorithms to their limits.
We found that CHOMP requires a large collision weight to navigate around the thin walls present in the \textit{plane} worlds. However, such a high value decimates performance in the \textit{sphere box} worlds, as the more `bulky' obstacles result in a very strong obstacle gradient, leading to unstable behavior that pushes CHOMP to the edges of, or even outside of, the world.
The baseline method and our neural network-based extension methods achieve surprisingly high success rates - especially considering the fact that none of these methods has access to a map. 

\subsection{Real-World Scenes}
While using very densely occupied simulated worlds gives us an idea of the relative performance of our algorithms, the true test of any learning-based system is how well it generalizes to real problems.
We evaluate several variants of proposed reactive planners as shown in the previous section (Baseline, FFN, and RNN and ablations) against each other, as well as CHOMP (CHMP40) and RRT*~\cite{karaman2011sampling}.

\begin{figure*}[t]
  \centering
  \begin{subfigure}{.5\linewidth}
    \centering
    \captionsetup{width=.95\linewidth}%
    \includegraphics[trim={100px 200px 400px 0px}, clip, width=.98\linewidth]{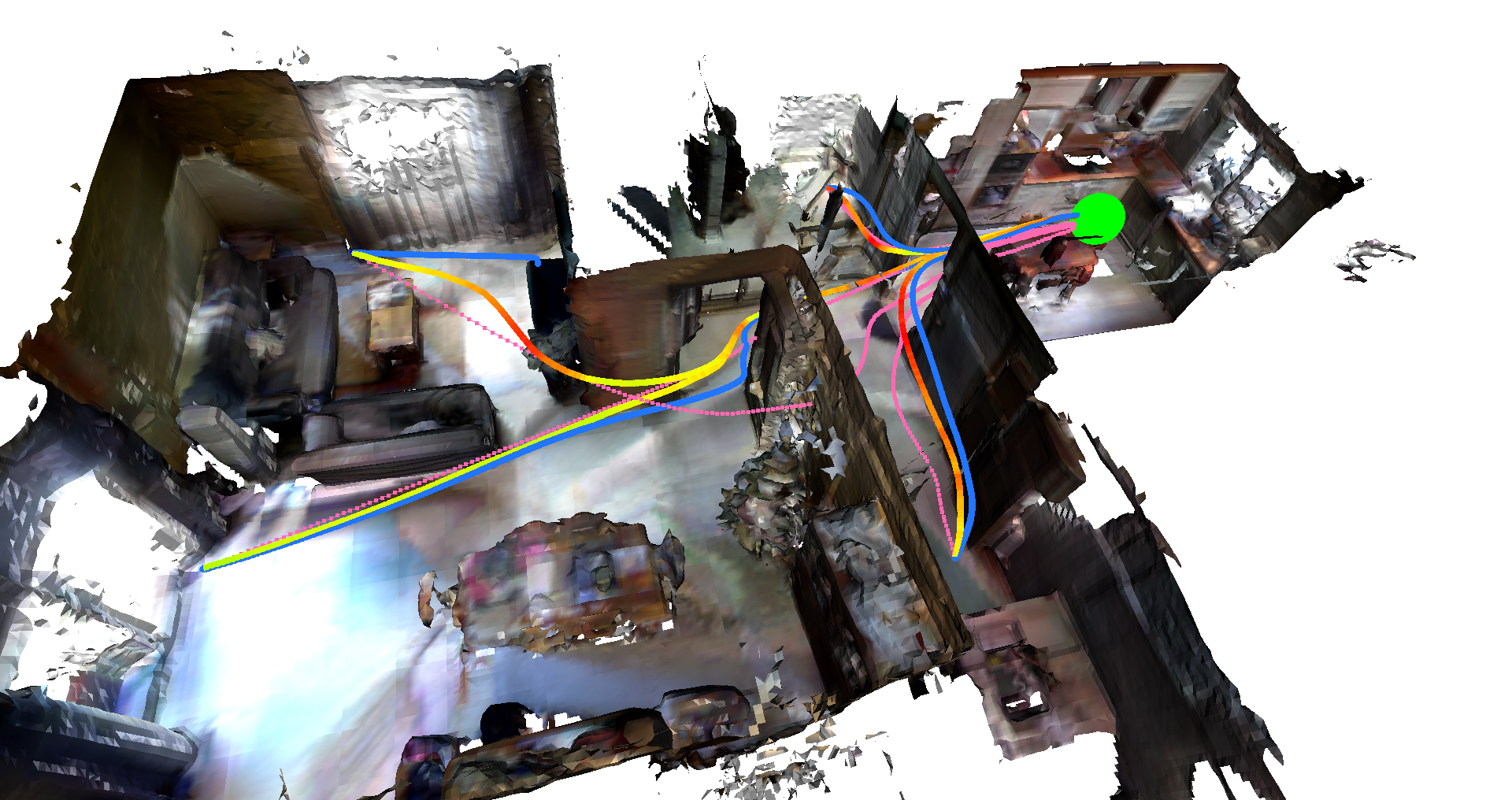}
    \caption{World from the SUN3D dataset~\cite{xiao2013sun3d}.}
    \label{fig:real_world_sun3d}
  \end{subfigure}%
  \begin{subfigure}{.5\linewidth}
    \centering 
    \captionsetup{width=.95\linewidth}
    \includegraphics[trim={400px 200px 100px 0px}, clip, width=.98\linewidth]{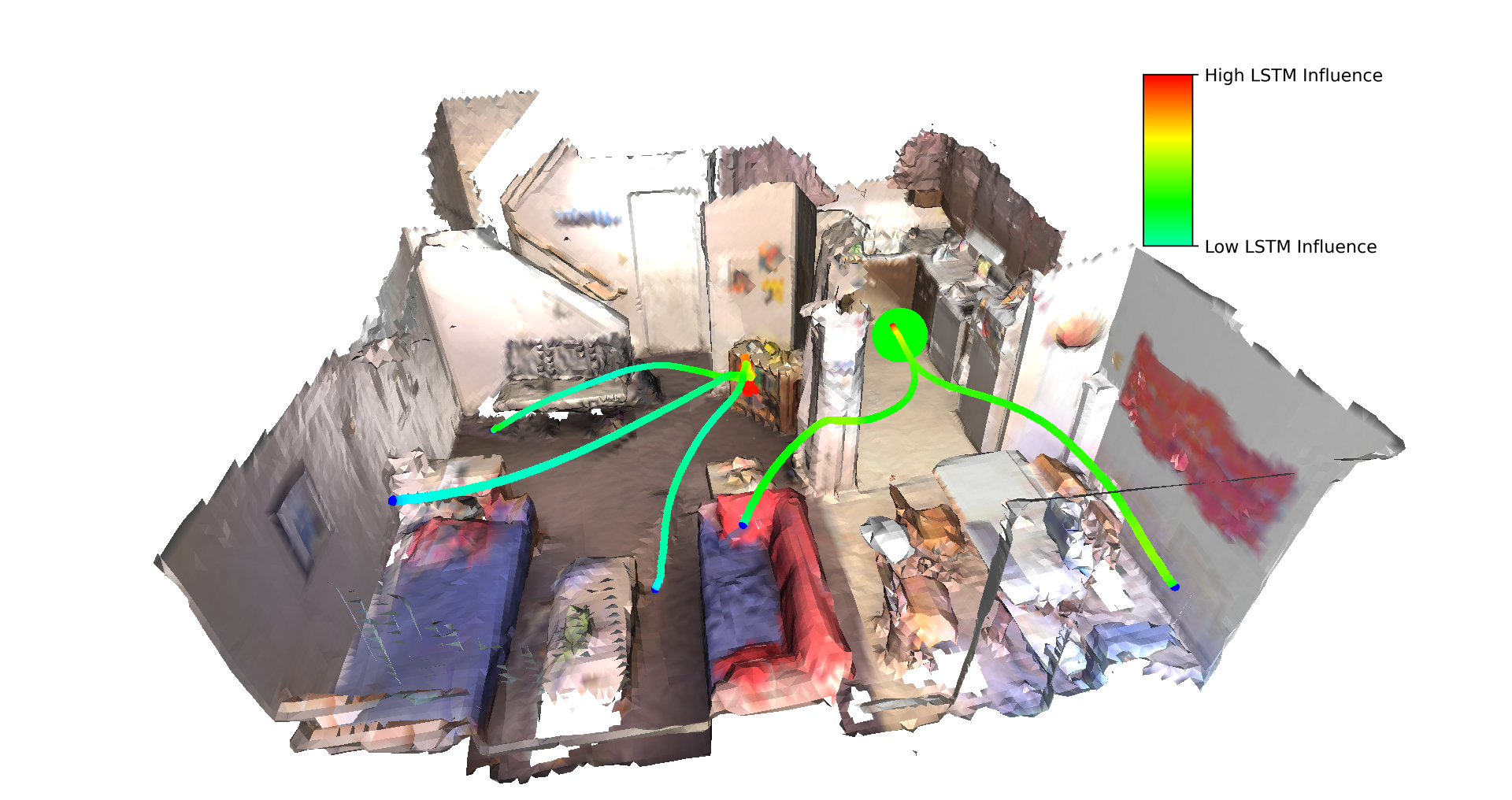}
    \caption[width=.95\linewidth]{World from the BundleFusion dataset~\cite{dai2017bundlefusion}.}
    \label{fig:real_world_bundle_apt0}
  \end{subfigure}%
  \caption{Navigation results on real-world data examples. The baseline shown in blue, CHOMP shown in pink, and the \ac{RNN} shown with the gradient color that shows the \ac{LSTM} level of influence. The learned system is transferred zero-shot to these environments, after being trained on the synthetically generated plane and sphere-box worlds. In (a) the \ac{RNN} is able to navigate past a local minimum in which the baseline gets stuck, and is also more efficient in general when navigating through dense geometry such as doorways. CHOMP is unable to deal with the longer trajectories through multiple doorways. In (b) the \ac{RNN} gets stuck in a larger local minimum.\vspace{-10pt}}
  \label{fig:real_world}
\end{figure*}

We use three datasets: a 120-obstacle variant of the generated \textit{sphere box} world (SB 120), the \textit{home\_at\_scan1\_2013\_jan\_1} sequence of SUN3D \cite{xiao2013sun3d}, shown in Figure \ref{fig:real_world_sun3d}, and the \textit{apt0} sequence of BundleFusion \cite{dai2017bundlefusion}, shown in \Cref{fig:real_world_bundle_apt0}.
We additionally do an ablation on the network size by introducing sFFN, which is a smaller version of the FFN with all learned layers with size $\geq 128$ halved.
The results are shown in \Cref{tab:stats}.

We show that our learned methods are able to zero-shot generalize to real environments, despite never having seen anything similar during training time. It is also again important to emphasize that CHOMP and RRT* have access to the complete map during planning time, while all the reactive methods (Base, sFFN, FFN, and RNN) only have access to the sensor rays at each planning timestep. As a result, our reactive methods also have a much smaller memory footprint: requiring only the ray information to be stored, not a complete volumetric reconstruction.

Another interesting result is that while all learned methods outperform the reactive baseline, the RNN only shows improved performance in simulation, not on the real-world datasets.
We hypothesize that the RNN ``overfits'' more to the distribution of environments used during training, as the effective input space of an RNN is much larger. Additionally, the smaller sFFN performs worse than the FFN on the synthetically generated world, but has roughly similar performance on the real world datasets. Similar to the RNN, we believe that the smaller network size may lead to better generalization to completely different environments.
As in previous evaluations, all learned methods outperform the CHOMP local planner. RRT*, as a global planning method with a much larger time budget, has the highest success rate -- but again, both methods have access to the full map while the reactive methods only have instantaneous sensor rays.

Another important aspect is that CHOMP and RRT* only provide output once the solver reaches a stopping criterion and a full trajectory is created, whereas the reactive methods simply need to evaluate the next-best acceleration at each time step.
The individual query time highlights this contrast, as the reactive methods can be queried orders of magnitude faster than CHOMP and RRT*. There is additional room for optimization: we found that a version of the \ac{RNN} compiled using the TensorRT compiler runs inference at $2.7$ kHz on an Nvidia Jetson Orin NX, which would lead to a query time of only 0.4 \textit{ms}.

%
%
\begin{table}
\centering
\setlength\tabcolsep{4pt} 
\begin{tabular}{l|l|cccccc}
\vspace{-5pt}
\textbf{World } & \textbf{ Plnr } & \textbf{ Map } & \textbf{Suc} & \thead{\textbf{Len}} & \thead{\textbf{TTA}} & \thead{\textbf{QT}} & \thead{\textbf{Mem}} \\
& & & & \textit{[m]} & \textit{[ms]} & \textit{[ms]} & \textit{[MB]} \\
\hline\multirow{6}{*}{\textbf{SB 120}} & CHMP & X & 0.78 & 7.67 & 113 & 113 & 3.9 \\
 & RRT* & X & 0.98 & 7.54 & 100 & 100 & 3.9 \\
 & Base &  & 0.75 & 7.22 & 169 & 0.4 & 0.1 \\
 & sFFN &  & 0.86 & 7.21 & 893 & 2.6 & 3.4 \\
 & FFN &  & 0.88 & 7.13 & 939 & 2.8 & 7.8 \\
 & RNN &  & 0.91 & 7.17 & 1900 & 5.7 & 9.9 \\
\hline\multirow{6}{*}{\textbf{SUN3D}} & CHMP & X & 0.65 & 6.08 & 279 & 279 & 159 \\
 & RRT* & X & 0.95 & 5.67 & 100 & 100 & 165 \\
 & Base &  & 0.56 & 5.64 & 153 & 0.5 & 0.1 \\
 & sFFN &  & 0.69 & 5.56 & 718 & 2.6 & 3.4 \\
 & FFN &  & 0.68 & 5.54 & 760 & 2.8 & 7.8 \\
 & RNN &  & 0.68 & 5.55 & 1540 & 5.6 & 9.9 \\
\hline\multirow{6}{*}{\textbf{BF}} & CHMP & X & 0.36 & 4.08 & 178 & 178 & 82 \\
 & RRT* & X & 1.00 & 3.75 & 100 & 100 & 90 \\
 & Base &  & 0.60 & 3.63 & 81 & 0.5 & 0.1 \\
 & sFFN &  & 0.74 & 3.64 & 445 & 2.6 & 3.4 \\
 & FFN &  & 0.74 & 3.64 & 485 & 2.8 & 7.8 \\
 & RNN &  & 0.72 & 3.64 & 1011 & 5.9 & 9.9 \\
\end{tabular}
\caption{Planning statistics for the \ac{RMP} planners, CHOMP and RRT*. The numbers originate from 100 planning runs with randomized start and goal locations that are at least $3m$ apart and not in direct line of sight.  We compare the SUN3D and BundleFusion (BF) worlds depicted in Figure \ref{fig:real_world}, and a sphere box world with 120 obstacles (SB 120). The columns indicate whether the method requires a full map, the success rate, the length of the trajectory, the time to get the full answer (TTA), the query time (QT), and the memory usage. Length, time to answer, and query time are all calculated on the subset of planning runs where \textit{all} methods are succesful per world type, such that the aggregated numbers can be compared directly. The experiments were conducted on AMD EPYC 7742 (single threaded) and Nvidia GeForce RTX 2080 Ti hardware.\vspace{-10pt}}
\label{tab:stats}
\end{table}
%

\subsection{Robustness to Noise}
We have shown that our learned methods generalize to real environments, but do they also generalize to realistic sensor noise?
We evaluate the robustness of the systems by adding noise to the rays (given as input to both the network and the avoidance policies). We choose a multiplicative noise model, where a ray distance $d_r$ is converted to a noisy observation using a normal distribution $\mathcal{N}\left(\cdot \right)$
\begin{equation}
    d_r \cdot \left(1 + \mathcal{N}\left(0, \sigma_{n}^2\right) \right), 
\end{equation}
parameterized by the noise standard deviation $\sigma_{n}$. This is done independently for every ray. \Cref{tab:noise} shows the performance of the RNN on the SUN3D world with different noise levels. The progression of these numbers is similar for different world types; performance is constant up to $30\%$, and quickly drops off afterwards. This indicates that the system is extremely robust to noise levels up to $30\%$, which is far above the noise levels reached by modern distance sensors.

\section{Discussion} 
\label{sec:discussion}
We show that adding a neural network to a purely reactive planner helps navigation in very cluttered environments.
By training on a geodesic field on a fully known map, we give our learned planners a form of geometric intuition on how to escape local minima based only on current sensor rays.
The addition of an \ac{LSTM} component further increased the success rate. We hypothesize that some temporal consistency is important to avoid certain types of local minima. Adding an \ac{LSTM} component also comes with trade-offs - it generally makes training less efficient and generalization outside of the training distribution is slightly worse. 
%
%
\begin{table}[t!]
    \centering
    \begin{tabular}{l|l|r|r|r|r|r}
         &\textbf{Noise stddev} $\sigma_{n}$ & $0\%$ & $10\%$ & $20\%$ & $30\%$ & $40\%$ \\
         \hline
         \multirow{2}{*}{\textbf{FFN}}&{Sucess Rate} & 0.68 & 0.68 & 0.68 & 0.68 & 0.33 \\
         &{Trajectory Length} \textit{[m]} & 4.80 & 4.81 & 4.80 & 4.80 & 5.12 \\
         \hline
         \multirow{2}{*}{\textbf{RNN}}&{Success Rate} & 0.68 & 0.69 & 0.69 & 0.69 & 0.40 \\
         &{Trajectory Length} \textit{[m]} & 4.81 & 4.81 & 4.81 & 4.81 & 5.01 \\
       
    \end{tabular}
    \caption{Success rate and trajectory lengths for different levels of noise for the FFN and RNN planner on the SUN3D worlds. The length is calculated on the subset of planning runs where \textit{all} methods in this table are successful, meaning that the aggregated lengths are different than in Table \ref{tab:stats}.\vspace{-10pt}}
    \label{tab:noise}
\end{table}
%

We used the geodesic field as a training supervision signal, as it acts a proxy for a global planner: pointing in the direction of the goal around obstacles throughout the map. However, the geodesic field is constructed such that it always points along the geodesic - the shortest possible path. The drawbacks of this are shown in Figure \ref{fig:rnn_stuck}, where the geodesic field is pointing in the direction of a very narrow opening above the wall, which might not be the ideal path for a real robot to traverse. While path length is often an important metric, it may not be the ultimate objective. Depending on the map, robot and scenario, it is sometimes better to take a longer path that fulfills other desired properties, e.g. avoiding narrow geometries or using fewer turns.
Especially in reactive navigation, finding the shortest path is not necessarily the ultimate objective; \textit{progress} towards the goal, by going towards large perceived open spaces, is often more useful. 

Another interesting aspect is the multi-modality of the navigation network output. In many situations there are multiple viable next directions that can be taken, which motivates future research into how to best exploit such output.

However, the results of this work provide meaningful and important insight into the difficulty and nature of reactive navigation in potentially cluttered 3D scenes \textit{in general}. Coming back to our introductory question -- when do we need a map? -- we here present evidence that for many realistic \textit{navigation} use-cases, reactive navigation combined with a higher-level geometric intuition suffices.
Considering a full robotic system, especially floating-base mobile robots where drift-free state estimation can be difficult to guarantee, the use of a pure reactive method greatly simplifies the operational complexity and state estimation quality requirements. By using a purely reactive navigation system that only needs the current sensor input to navigate safely, robot designers can use less accurate odometry sources, have fewer requirements on time synchronization between sensors, and reduce overall computational complexity of their systems. Furthermore, the navigation policies themselves are extremely robust to noise on the input sensor data, hopefully bringing safety even with lower-cost sensors. 

In this work, we used a generic ray-casting interface to a mapping system to mimic reactive sensor navigation. As was shown for the baseline method \cite{pantic2023obstacle}, the ray-casting interface closely resembles LiDAR data. In future work, we plan to study the effects of using LiDAR data for navigation.
\section{Conclusion} 
In this paper we presented two neural-network-based navigation methods, that when combined with a purely reactive safety layer, enable navigation through very densely cluttered 3D worlds using only local sensor data and without a map. Our system outperforms other local, well-known methods and is trained in a fully self-supervised fashion in auto-generated worlds. Additionally, it is capable of zero-shot transfer to real 3D environments, and has high robustness to noise. The modular architecture facilitates the seamless combination of the navigation stack with any other task formulated as an \ac{RMP}, and enables introspection to gain intuition about \textit{how} the learned components exert their influence. We exploit the introspectability of the presented system for understanding the challenges and nature of local navigation in 3D spaces.
The ability of the system to find its way through extremely cluttered maps with only local data is surprising and highly relevant for practical applications, where robust traversal of our cluttered and semi-structured world with minimal system requirements is highly important.
\bibliographystyle{unsrtnat}
\bibliography{references}

\begin{thebibliography}{22}
\providecommand{\natexlab}[1]{#1}
\providecommand{\url}[1]{\texttt{#1}}
\expandafter\ifx\csname urlstyle\endcsname\relax
  \providecommand{\doi}[1]{doi: #1}\else
  \providecommand{\doi}{doi: \begingroup \urlstyle{rm}\Url}\fi

\bibitem[Dai et~al.(2017)Dai, Nie\ss{}ner, Zollh\"{o}fer, Izadi, and Theobalt]{dai2017bundlefusion}
Angela Dai, Matthias Nie\ss{}ner, Michael Zollh\"{o}fer, Shahram Izadi, and Christian Theobalt.
\newblock Bundlefusion: Real-time globally consistent 3d reconstruction using on-the-fly surface reintegration.
\newblock \emph{ACM Trans. Graph.}, 2017.
\newblock ISSN 0730-0301.
\newblock \doi{10.1145/3072959.3054739}.

\bibitem[Khatib(1986)]{khatib1986potential}
Oussama Khatib.
\newblock The potential field approach and operational space formulation in robot control.
\newblock In \emph{Adaptive and Learning Systems: Theory and Applications}. Springer, 1986.

\bibitem[Quinlan and Khatib(1993)]{quinlan1993elastic}
Sean Quinlan and Oussama Khatib.
\newblock Elastic bands: Connecting path planning and control.
\newblock In \emph{IEEE International Conference on Robotics and Automation (ICRA)}. IEEE, 1993.

\bibitem[Ratliff et~al.(2009)Ratliff, Zucker, Bagnell, and Srinivasa]{ratliff2009CHOMP}
Nathan Ratliff, Matt Zucker, J.~Andrew Bagnell, and Siddhartha Srinivasa.
\newblock Chomp: Gradient optimization techniques for efficient motion planning.
\newblock In \emph{IEEE International Conference on Robotics and Automation (ICRA)}, 2009.
\newblock \doi{10.1109/ROBOT.2009.5152817}.

\bibitem[Oleynikova et~al.(2018)Oleynikova, Taylor, Siegwart, and Nieto]{oleynikova2018safe}
Helen Oleynikova, Zachary Taylor, Roland Siegwart, and Juan Nieto.
\newblock Safe local exploration for replanning in cluttered unknown environments for microaerial vehicles.
\newblock \emph{IEEE Robotics and Automation Letters (RA-L)}, 2018.

\bibitem[Karaman and Frazzoli(2011)]{karaman2011sampling}
Sertac Karaman and Emilio Frazzoli.
\newblock Sampling-based algorithms for optimal motion planning.
\newblock \emph{The International Journal of Robotics Research}, 2011.
\newblock \doi{10.1177/0278364911406761}.

\bibitem[Montano and Asensio(1997)]{montano1997real}
Luis Montano and Jos{\'e}~R Asensio.
\newblock Real-time robot navigation in unstructured environments using a 3d laser rangefinder.
\newblock In \emph{IEEE/RSJ International Conference on Intelligent Robot and Systems (IROS)}. IEEE, 1997.

\bibitem[Ratliff et~al.(2018)Ratliff, Issac, Kappler, Birchfield, and Fox]{ratliff2018riemannian}
Nathan~D Ratliff, Jan Issac, Daniel Kappler, Stan Birchfield, and Dieter Fox.
\newblock Riemannian motion policies.
\newblock \emph{arXiv preprint arXiv:1801.02854}, 2018.

\bibitem[Pantic et~al.(2023)Pantic, Meijer, Bähnemann, Alatur, Andersson, Cadena, Siegwart, and Ott]{pantic2023obstacle}
Michael Pantic, Isar Meijer, Rik Bähnemann, Nikhilesh Alatur, Olov Andersson, Cesar Cadena, Roland Siegwart, and Lionel Ott.
\newblock Obstacle avoidance using raycasting and riemannian motion policies at khz rates for mavs.
\newblock In \emph{IEEE International Conference on Robotics and Automation (ICRA)}, 2023.
\newblock \doi{10.1109/ICRA48891.2023.10161365}.

\bibitem[Mattamala et~al.(2022)Mattamala, Chebrolu, and Fallon]{mattamala2022efficient}
Matias Mattamala, Nived Chebrolu, and Maurice Fallon.
\newblock An efficient locally reactive controller for safe navigation in visual teach and repeat missions.
\newblock \emph{IEEE Robotics and Automation Letters}, 2022.

\bibitem[Mirowski et~al.(2016)Mirowski, Pascanu, Viola, Soyer, Ballard, Banino, Denil, Goroshin, Sifre, Kavukcuoglu, et~al.]{mirowski2017learning}
Piotr Mirowski, Razvan Pascanu, Fabio Viola, Hubert Soyer, Andy Ballard, Andrea Banino, Misha Denil, Ross Goroshin, Laurent Sifre, Koray Kavukcuoglu, et~al.
\newblock Learning to navigate in complex environments.
\newblock In \emph{International Conference on Learning Representations (ICRL)}, 2016.

\bibitem[Song et~al.(2023)Song, Shi, Penicka, and Scaramuzza]{song2023learning}
Yunlong Song, Kexin Shi, Robert Penicka, and Davide Scaramuzza.
\newblock Learning perception-aware agile flight in cluttered environments.
\newblock In \emph{IEEE International Conference on Robotics and Automation (ICRA)}. IEEE, 2023.

\bibitem[Ross et~al.(2013)Ross, Melik-Barkhudarov, Shankar, Wendel, Dey, Bagnell, and Hebert]{ross2013learning}
St{\'e}phane Ross, Narek Melik-Barkhudarov, Kumar~Shaurya Shankar, Andreas Wendel, Debadeepta Dey, J~Andrew Bagnell, and Martial Hebert.
\newblock Learning monocular reactive uav control in cluttered natural environments.
\newblock In \emph{IEEE International Conference on Robotics and Automation (ICRA)}. IEEE, 2013.

\bibitem[Ross et~al.(2011)Ross, Gordon, and Bagnell]{ross2011reduction}
Stephane Ross, Geoffrey Gordon, and Drew Bagnell.
\newblock A reduction of imitation learning and structured prediction to no-regret online learning.
\newblock In \emph{International Conference on Artificial Intelligence and Statistics}, Proceedings of Machine Learning Research. PMLR, 2011.

\bibitem[Loquercio et~al.(2021)Loquercio, Kaufmann, Ranftl, Müller, Koltun, and Scaramuzza]{loquercio2021learning}
Antonio Loquercio, Elia Kaufmann, René Ranftl, Matthias Müller, Vladlen Koltun, and Davide Scaramuzza.
\newblock Learning high-speed flight in the wild.
\newblock \emph{Science Robotics}, 2021.
\newblock \doi{10.1126/scirobotics.abg5810}.

\bibitem[Tai et~al.(2017)Tai, Paolo, and Liu]{tai2017virtual}
Lei Tai, Giuseppe Paolo, and Ming Liu.
\newblock Virtual-to-real deep reinforcement learning: Continuous control of mobile robots for mapless navigation.
\newblock In \emph{IEEE/RSJ International Conference on Intelligent Robots and Systems (IROS)}, 2017.
\newblock \doi{10.1109/IROS.2017.8202134}.

\bibitem[Pfeiffer et~al.(2018)Pfeiffer, Shukla, Turchetta, Cadena, Krause, Siegwart, and Nieto]{pfeiffer2018reinforced}
Mark Pfeiffer, Samarth Shukla, Matteo Turchetta, Cesar Cadena, Andreas Krause, Roland Siegwart, and Juan Nieto.
\newblock Reinforced imitation: Sample efficient deep reinforcement learning for mapless navigation by leveraging prior demonstrations.
\newblock \emph{IEEE Robotics and Automation Letters (RA-L)}, 2018.
\newblock \doi{10.1109/LRA.2018.2869644}.

\bibitem[Zhang et~al.(2017)Zhang, Tai, Liu, Boedecker, and Burgard]{zhang2017neural}
Jingwei Zhang, Lei Tai, Ming Liu, Joschka Boedecker, and Wolfram Burgard.
\newblock Neural slam: Learning to explore with external memory.
\newblock \emph{arXiv preprint arXiv:1706.09520}, 2017.

\bibitem[Millane et~al.(2023)Millane, Oleynikova, Wirbel, Steiner, Ramasamy, Tingdahl, and Siegwart]{millane2023nvblox}
Alexander Millane, Helen Oleynikova, Emilie Wirbel, Remo Steiner, Vikram Ramasamy, David Tingdahl, and Roland Siegwart.
\newblock nvblox: Gpu-accelerated incremental signed distance field mapping.
\newblock \emph{arXiv preprint arXiv:2311.00626}, 2023.

\bibitem[Halton(1964)]{halton1964algorithm}
John~H Halton.
\newblock Algorithm 247: Radical-inverse quasi-random point sequence.
\newblock \emph{Communications of the ACM}, 1964.

\bibitem[Sethian(1996)]{sethian1996fast}
James~A Sethian.
\newblock A fast marching level set method for monotonically advancing fronts.
\newblock \emph{Proceedings of the National Academy of Sciences}, 1996.

\bibitem[Xiao et~al.(2013)Xiao, Owens, and Torralba]{xiao2013sun3d}
Jianxiong Xiao, Andrew Owens, and Antonio Torralba.
\newblock Sun3d: A database of big spaces reconstructed using sfm and object labels.
\newblock In \emph{IEEE International Conference on Computer Vision (ICCV)}, 2013.

\end{thebibliography}
\end{document}